\documentclass[runningheads]{llncs}
\usepackage{wrapfig}
\usepackage{graphicx}
\usepackage{url}
\usepackage{graphicx}
\usepackage{tabularray}
\usepackage{tikz}
\usepackage{comment}
\usepackage{amsmath,amssymb} 
\usepackage{color}
\usepackage{ulem}
\usepackage{graphicx}
\usepackage{subcaption}
\usepackage{bbding}
\usepackage[accsupp]{axessibility}  
\usepackage{amsmath}
\usepackage[misc]{ifsym}
\usepackage[colorlinks,hyperindex,breaklinks]{hyperref}
\usepackage{multirow}
\usepackage{booktabs}
\usepackage[T1]{fontenc}
\newcommand{\etal}{\textit{et al}.}

\begin{document}
\title{Revisiting Deep Ensemble Uncertainty for Enhanced Medical Anomaly Detection}
\author{Yi Gu$^{\dag 1}$, Yi Lin$^{\dag 1}$, Kwang-Ting Cheng$^2$, and Hao Chen\textsuperscript{\Letter}$^{1,3,4}$}
\institute{$^{1}$Department of Computer Science and Engineering, HKUST, Hong Kong, China\\ 
$^{2}$Department of Electronic and Computer Engineering, HKUST, Hong Kong, China\\
$^{3}$Department of Chemical and Biological Engineering, HKUST, Hong Kong, China\\ 
$^{4}$HKUST Shenzhen-Hong Kong Collaborative Innovation Research Institute, Futian, Shenzhen, China\\
\email{yguax@connect.ust.hk, jhc@cse.ust.hk} 
}
\titlerunning{D2UE}
\authorrunning{Y. Gu and Y. Lin, \etal}
\maketitle       
\def\thefootnote{$\dag$}\footnotetext{Equal contribution; \Letter~corresponding author.}
\begin{abstract}

Medical anomaly detection (AD) is crucial in pathological identification and localization. 
Current methods typically rely on uncertainty estimation in deep ensembles to detect anomalies, assuming that ensemble learners should agree on normal samples while exhibiting disagreement on unseen anomalies in the output space. 
However, these methods may suffer from inadequate disagreement on anomalies or diminished agreement on normal samples. 
To tackle these issues, we propose D2UE, a Diversified Dual-space Uncertainty Estimation framework for medical anomaly detection. 
To effectively balance agreement and disagreement for anomaly detection, we propose Redundancy-Aware Repulsion (RAR), which uses a similarity kernel that remains invariant to both isotropic scaling and orthogonal transformations, explicitly promoting diversity in learners' feature space. 
Moreover, to accentuate anomalous regions, we develop Dual-Space Uncertainty (DSU), which utilizes the ensemble's uncertainty in input and output spaces.
In input space, we first calculate gradients of reconstruction error with respect to input images.
The gradients are then integrated with reconstruction outputs to estimate uncertainty for inputs, enabling effective anomaly discrimination even when output space disagreement is minimal.
We conduct a comprehensive evaluation of five medical benchmarks with different backbones. Experimental results demonstrate the superiority of our method to state-of-the-art methods and the effectiveness of each component in our framework. 
Our code is available at \url{https://github.com/Rubiscol/D2UE}.

\end{abstract}

\section{Introduction}
Anomaly detection (AD) is an essential task in medical image analysis, encompassing early detection of medical diseases~\cite{Ganomaly,IGD} and pathological localization~\cite{DBLP:journals/mia/fanogan}. The primary objective of visual medical AD is to identify images containing diseases and pinpoint anomalous pixels within them. However, obtaining a sufficient number of anomalous samples that cover the vast spectrum of disease types can be challenging, as these samples often require specialized annotations~\cite{DDAD}. Consequently, AD tasks are often formulated as one-class classification problem, wherein only normal data is utilized for model training~\cite{cai2024medianomaly}.

Prevailing approaches mainly focus on reconstruction-based anomaly detection employing Autoencoders~\cite{memAE,AEU} or Generative Adversarial Networks~\cite{DBLP:journals/mia/fanogan,Ganomaly}. 
These methods endeavor to maximize the likelihood of normal samples derived from training data. During inference, anomalies are detected based on per-pixel reconstruction error or model probability distribution. Nevertheless, these methods are limited by imprecise reconstructions or poorly calibrated likelihoods~\cite{cai2024rethinking}. 

To circumvent a direct estimation of normal probability distributions, an alternative framework leveraging deep ensembles' uncertainty has emerged. This framework comprises multiple learners that perform self-supervised tasks~\cite{DDAD,DBLP:conf/miccai/AMAE} or acquires surrogate labels through a pretrained encoder~\cite{ST-AD,MR-AD}. Typically, learners undergo randomized training with distinct weight initializations~\cite{DDAD} or Monte-Carlo dropout~\cite{DBLP:conf/icml/Dropout}. The underlying hypothesis posits that diverse learners should agree on normality while disagreeing on unseen anomalies in the output space.
\begin{figure}[t]
  \centering
  \includegraphics[width=\linewidth]{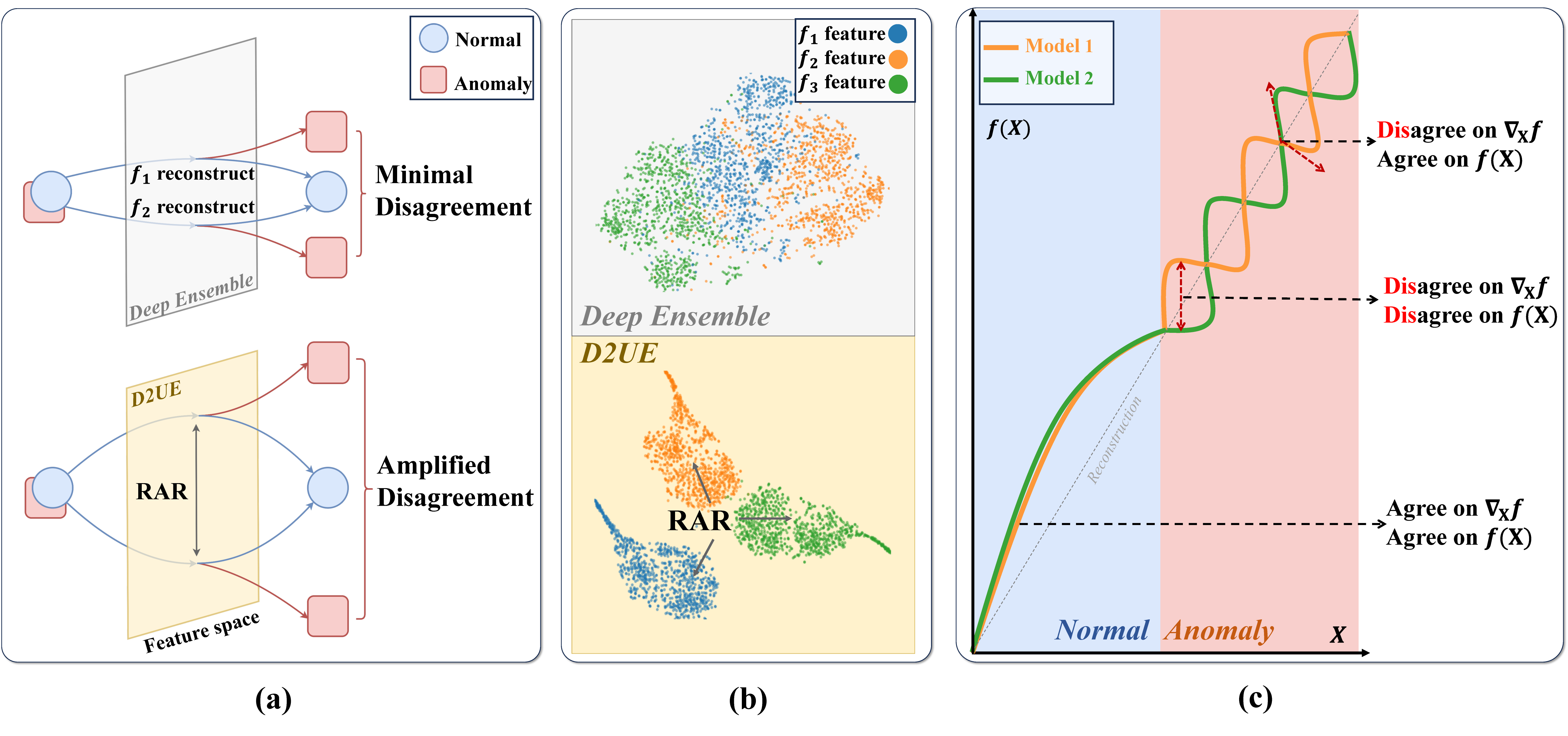}
  \caption{\textbf{(a)}: An illustration of redundancy-aware repulsion (RAR). Disagreement on anomalies is amplified between different learners’ feature spaces, while normal input converges to similar reconstructions guided by reconstruction training. \textbf{(b)}: A t-SNE~\cite{tSNE} plot of feature spaces from three learners on the anomaly. Feature spaces are pushed away by RAR during training. \textbf{(c)}: An illustration of dual-space uncertainty (DSU) in 1D regression with two learners. Utilizing output space uncertainty fails to differentiate the anomaly at the upper point. In comparison, DSU utilizes the disagreement on $\nabla_{X}{f}$ to detect such anomalies.}
  \label{fig:intro}
\end{figure}

However, balancing the trade-off between agreement and disagreement is challenging. Randomized training may not guarantee sufficient disagreement on anomalies, as learners in ensemble learning inherently tend to adopt the simplest decision boundary~\cite{DBLP:conf/icml/SB_1}. This phenomenon, known as simplicity bias~\cite{DBLP:conf/nips/SB_2}, inhibits learners' diversity and subsequently results in minimal disagreements on anomaly outputs~\cite{lin2023nuclei}. To address simplicity bias, previous methods attempted to induce repulsion among learners in output space~\cite{DBLP:conf/iclr/AgreetoDisagree} or weight space~\cite{d2021repulsive}. Nevertheless, these approaches may culminate in either underfitting of individual models~\cite{Underfit} or neural network redundancy~\cite{lin2024lenas}, where models possess distinct weights yet output the same~\cite{Weight_sym}. Consequently, learners' agreement on normal samples would be compromised.

In this paper, we propose a novel ensemble-based uncertainty estimation framework for medical anomaly detection called Diversified Dual-space Uncertainty Estimation (\textbf{D2UE}). To enhance learners' disagreement on anomalies, we introduce a Redundancy-Aware Repulsion (\textbf{RAR}), which encourages learners to reconstruct training samples from more diversified feature spaces. To promote this diversification without succumbing to neural network redundancy, RAR regulates ensemble training using a similarity kernel invariant to both isotropic scaling and orthogonal transformation. During inference, disagreement on anomalies is amplified between different learners' feature spaces (see Fig.~\ref{fig:intro}(a)). Unlike output space repulsion, feature space repulsion does not result in underfitting for normal samples in output space. Consequently, normal features converge to similar reconstructions guided by reconstruction training. Moreover, to emphasize anomalous regions, we develop a Dual-Space Uncertainty (\textbf{DSU}) that combines uncertainties in both input and out spaces. In input space, we calculate gradients of the reconstruction error with respect to inputs, which are further combined with outputs to estimate the final uncertainty. DSU discriminates anomalies through input space disagreement even if learners exhibit minimal disagreement in output space (see Fig.~\ref{fig:intro}(c)). Our primary contributions are as follows:

\begin{itemize} 
    \item We address medical anomaly detection from an uncertainty estimation perspective. We undertake a pioneering exploration of diversity in deep ensembles' uncertainty and propose D2UE, a novel Diversified Dual-space Uncertainty Estimation approach for medical anomaly detection. 
    \item We propose Redundancy-Aware Repulsion (RAR) to strike an effective balance between agreement and disagreement, thereby enhancing anomaly detection accuracy. 
    \item We design a Dual-Space Uncertainty (DSU) to emphasize anomalous regions, particularly when anomalies exhibit minimal disagreement in output space. 
    \item We conduct comprehensive experiments on five medical benchmarks with different backbones. Experimental results demonstrate the superiority of our method to state-of-the-art methods and the effectiveness of each component.\end{itemize}              
\label{introduction}
\section{Method}
\label{sec:overview}
The proposed Diversified Dual-space Uncertainty Estimation (D2UE) framework consists of $N$ learners $f(\cdot)$, each possessing identical Autoencoder architectures. 
Anomaly detection is achieved through uncertainty estimation, where learners should agree on normal samples while disagreeing on anomalies. 
To this end, we propose redundancy-aware repulsion (RAR) to enhance learners' disagreement on anomalies while maintaining agreement on normal samples.
Moreover, to further emphasize anomalous regions, dual-space uncertainty (DSU) is designed to combine uncertainties in output and input spaces during inference. 
In the following, we will detail each component.
\begin{figure}[t]
  \centering
  \includegraphics[width=\linewidth]{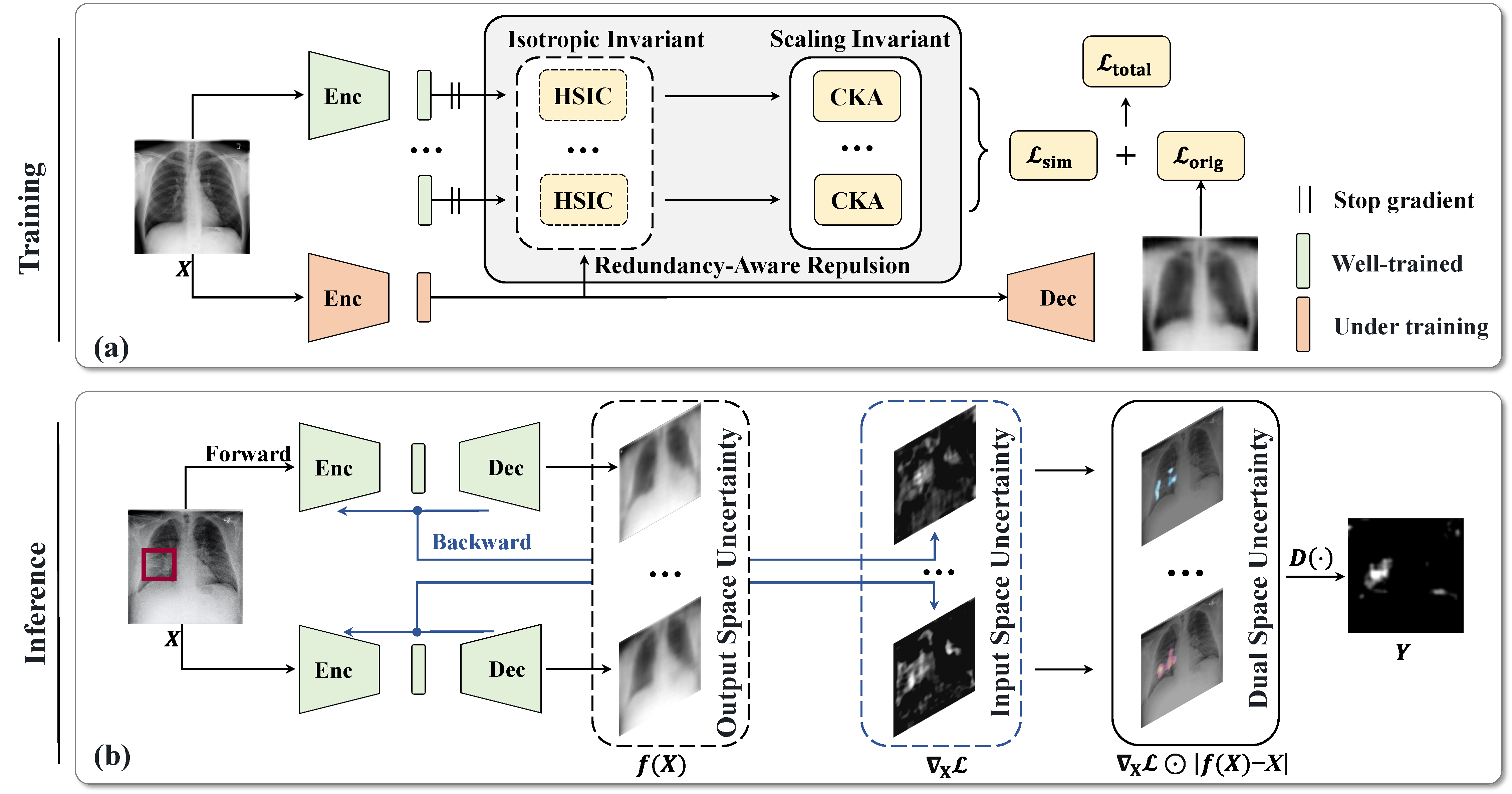}
  \caption{Overview of D2UE. In the training stage, the redundancy-aware repulsion (RAR) module amplifies the diversity of different models with both isotropic and scaling invariance. In the inference stage, the dual-space uncertainty is calculated, utilizing both $f(X)$ in the output space and $\nabla_{X}\mathcal{L}$ in the input space.}
  \label{fig:overview}
\end{figure}

\noindent\textbf{Redundancy-Aware Repulsion for Feature Space.}\label{sec:constraint} Existing methods mainly train ensemble learners under the reconstruction loss $\mathcal{L}_{\text{orig}}$, such as mean square error loss, without any repulsion encouraging constraint. To encourage learners' disagreement on anomalies, our approach incorporates a similarity constraint loss $\mathcal{L}_{\text{sim}}$ that induces repulsion in feature space. As depicted in Fig.~\ref{fig:overview}, learners undergo sequential training and the learner under training is denoted as $f_n$. The $\mathcal{L}_\text{sim}$ of $f_n$, except for 0 in the first one, is optimized to minimize the Centered Kernel Alignment (CKA similarity)~\cite{CKA} between its feature vector $Q^n$ and well-trained learner's feature $P^j$ in total $n-1$ trained learners: 
\begin{align}\label{eq:1}
\mathcal{L}_{\text{sim}}= \frac{1}{n-1} \sum_{j=1}^{n-1} \text{CKA}(P^j,Q^n) 
\end{align}
\begin{align}
\label{eq:2}
    \text{CKA}(P,Q)=\frac{\text{HSIC}(K,L)}{\sqrt{\text{HSIC}(K,K)} \sqrt{\text{HSIC}(L,L)}}
\end{align}
In Eq.~(\ref{eq:2}), $K=PP^T$ and $L=QQ^T$, HSIC is Hilbert-Schmidt Independence Criterion~\cite{DBLP:conf/nips/HSIC} measuring the independence between variables:
\begin{align}
\label{eq:3}
    \text{HSIC}(K,L)=\frac{1}{(r-1)^2} tr (KHLH)
\end{align}
where $r$ is the rank of $K$ and $L$, $H$ is the centering matrix, and $tr$ is the matrix trace.
In the following, we will elaborate the motivation behind our design. 

First, it is crucial to establish an optimization objective that explicitly encourages repulsion in learners' feature spaces. To this end, we endow a more straightforward target to the training of a learner: to reconstruct normal samples using $\mathcal{L}_{\text{orig}}$ through more different paths supervised by all well-trained learners via $\mathcal{L}_{\text{sim}}$. By passing anomalous samples through more diversified feature spaces, learners are encouraged to exhibit more significant disagreements while maintaining consistency on normal samples during inference.

\begin{figure}[t]
  \centering
  \includegraphics[width=\linewidth]{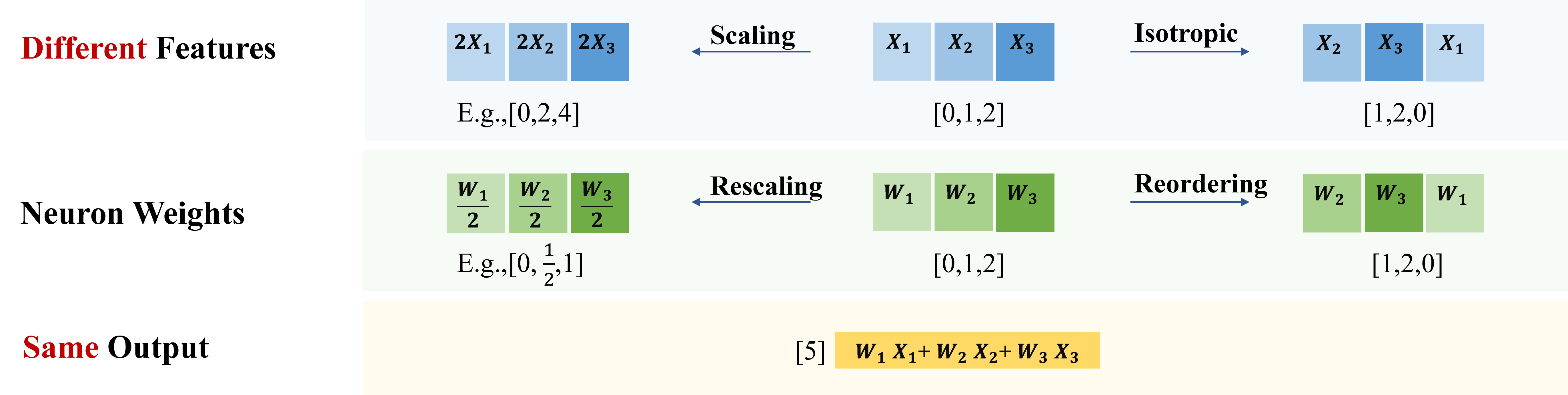}
  \caption{An illustration of neural network redundancy. Different features may output the same by weight re-scaling or spatial reordering.}
  \label{fig:redundancy}
\end{figure}

Second, it is essential to identify a suitable similarity kernel $S(\cdot)$ that eliminates neural network redundancy. To achieve this, we establish two properties for $S(\cdot)$: 1. \textbf{isotropic scaling invariance}: $S(P,Q)=S(\alpha P,\beta Q)$ for $\alpha, \beta \in \mathbb{R}^+$. 2. \textbf{orthogonal transformation invariance}: $S(P,Q)=S(PU,QV)$ for $U,V\in$ orthogonal transformation. Scaling invariance implies that neurons cannot deceive well-trained neurons into perceiving different features merely by re-scaling their weights during training. Similarly, orthogonal invariance prevents neurons from deceiving well-trained neurons into perceiving different features merely by spatial reordering during training. As illustrated in Fig.~\ref{fig:redundancy}, different features may still output the same by weight re-scaling or reordering. Therefore, both isotropic and orthogonal invariance are necessary to effectively promote feature space diversity without succumbing to neural network redundancy.

In Eq.~(\ref{eq:3}), $K=PP^T$ and $L=QQ^T$ ensure orthogonal invariance since $(PU)(PU)^T=PP^T$. Normalized term $\sqrt{\text{HSIC}(K,K)} \sqrt{\text{HSIC}(L,L)}$ ensures scaling invariance in $\mathcal{L}_{\text{sim}}$. Finally, the total loss $\mathcal{L}_{\text{total}}$ is formulated as follows:
\begin{equation}\label{eq:4}
\mathcal{L}_{\text{total}}=\underbrace{\mathcal{L}_\text{orig}}_{\text{agree on normal}} + \underbrace{\lambda\mathcal{L}_{\text{sim}_{_{}}}}_{\text{disagree on anomaly}}
\end{equation} where $\lambda$ controls the strength of the repulsion to the overall loss function\footnote{Ablation studies of $\lambda$ and the layer of $\mathcal{L}_{\text{sim}}$ are included in supplementary materials.}. 

\noindent\textbf{Dual-Space Uncertainty.} \label{sec:bd} To identify anomalous samples, an image $X$ is input to all learners to estimate uncertainty, generating a pixel-level anomaly score map $Y$. Previous methods calculated $Y$ based on models' outputs $f(X)$:
\begin{equation}\label{eq:5}
    Y=D(f_i({X})), i=1....N
\end{equation}wherein $D(\cdot)$ signifies the deviation function. However, relying solely on the uncertainty of $f(X)$ may fail to discriminate anomalies in some cases. In the context of the same reconstruction task, learners can sometimes output similar reconstructions even for anomalies, as depicted in Fig.~\ref{fig:intro}(c). 
Despite minimal disagreement in output space, a model can hold unique first-order derivatives $\nabla_Xf$ in input space~\cite{trinh2023input}. Typically, $\nabla_Xf$ is used to construct saliency maps and visually interpret models' divergent attention on input pixels~\cite{DBLP:journals/corr/SimonyanVZ13}. Inspired by this, we devise DSU that correlates output space uncertainty with input space uncertainty to better reveal learners' disagreement towards anomalies. Specifically, in input space, we calculate $\nabla_{X}\mathcal{L}$, the gradient of reconstruction error with respect to the input, as opposed to a large Jacobian matrix $\nabla_Xf$ in the reconstruction model. The $\nabla_{X}\mathcal{L}$ is further elementally multiplied with normalized output $| f({X})-X |$ to calculate the final uncertainty:

\begin{equation}\label{eq:6}
    Y=D(\nabla_{X}\mathcal{L}_i\odot| f_i({X})-X |), i=1...N
\end{equation} where $\odot$ symbolizes the element-wise multiplication. Consequently, even if distinct learners inadvertently agree in output space, the anomalous region can still be accentuated by input space disagreement.

\label{method}
\section{Experiment}
\textbf{Datasets and Evaluation metrics}. We conduct experiments on five medical datasets, encompassing various modalities such as chest X-ray, magnetic resonance imaging, and retinal fundus. Datasets include 1. RSNA: RSNA Pneumonia Detection Challenge dataset\footnote{https://www.kaggle.com/c/rsna-pneumonia-detection-challenge}. 2. VinDr-CXR: VinBigData Chest X-ray Abnormalities Detection dataset\footnote{https://www.kaggle.com/c/vinbigdata-chest-xray-abnormalities-detection}. 3. CXAD: Chest X-ray Anomaly Detection dataset~\cite{DDAD}. 4. Brain MRI: Brain Tumor MRI dataset\footnote{https://www.kaggle.com/datasets/masoudnickparvar/brain-tumor-mri-dataset}. 5. LAG: Large-scale Attention-based Glaucoma dataset~\cite{LAG}. To ensure consistency with other studies, we adhere to the split criteria outlined in~\cite{DBLP:conf/miccai/AMAE,DDAD,DBLP:journals/mia/DDAD-ASR} and details can be seen in supplementary materials. We adopt the area under the ROC curve (AUC) and average precision (AP) for image-level classification.

\noindent\textbf{Implementation Details}.
We optimize our model using the Adam optimizer with an initial learning rate of $5e-4$ for AE and MemAE, and $1e-3$ for AEU. The default batch size is set to 64 with the image size of $64\times64$. Each learner is randomly initialized and trained for 250 epochs. $\lambda$ is set to 1 in our experiment.

\begin{table}[t]
\centering
\caption{Comparison with SOTA methods. ``Reconstruction" refers to ``Reconstruction methods", ``Uncertainty" refers to ``Uncertainty estimates methods". The best result is highlighted in bold and the second best result is underline.}
\label{tab:comparison}
\resizebox{\columnwidth}{!}{
\begin{tabular}{clcccccccccccc}
\toprule
\multicolumn{2}{c}{\multirow{2}{*}{Methods}} & \multicolumn{2}{c}{RSNA} & \multicolumn{2}{c}{VinDr-CXR} & \multicolumn{2}{c}{CXAD} & \multicolumn{2}{c}{Brain  MRI} & \multicolumn{2}{c}{LAG} & \multicolumn{2}{c}{Average} \\
\cmidrule(lr){3-4} \cmidrule(lr){5-6} \cmidrule(lr){7-8} \cmidrule(lr){9-10} \cmidrule(lr){11-12} \cmidrule(lr){13-14} 
        && AUC & AP & AUC & AP & AUC & AP & AUC & AP & AUC & AP & AUC & AP   \\
\midrule

\multirow{6}{*}{\rotatebox{90}{Reconstruction}}    & AE          & 66.9                  & 66.1                   & 55.9                  & 60.3                  & 55.6                  & 59.6                  & 79.7                  & 71.9                  & 79.3                  & 76.1                  & 67.5    & 66.8 \\
        & MemAE~\cite{memAE}~      & 68.0                  & 67.1                   & 55.8                  & 59.8                  & 56.0                  & 60.0                  & 77.4                  & 70.0                  & 78.5                  & 74.9                  & 66.7    & 66.6 \\
        & AEU~\cite{AEU}         & 86.7                  & 84.7                   & 73.8                  & 72.8                  & 66.4                  & 66.9                  & 94.0                  & 89.0                  & 81.3                  & 78.9                  & \uline{80.4}    & \uline{78.5} \\
        & IGD~\cite{IGD}~        & 81.2                  & 78.0                   & 59.2                  & 58.7                  & 55.2                  & 57.6                  & 94.3                  & 90.6                  & 80.7                  & 75.3                  & 74.1    & 72.0 \\
        & f-AnoGAN~\cite{DBLP:journals/mia/fanogan}   & 79.8                  & 75.6                   & 76.3                  & \uline{74.8}         & 61.9                  & 67.3                  & 82.5                  & 74.3                  & \uline{84.2}         & 77.5                  & 76.9    & 73.9 \\
        & Ganomaly~\cite{Ganomaly}~   & 71.4                  & 69.1                   & 59.6                  & 60.3                  & 62.5                  & 63.0                  & 75.1                  & 69.7                  & 77.7                  & 75.7                  & 69.3    & 67.6 \\
\midrule
\multirow{6}{*}{\rotatebox{90}{Uncertainty}}    & DDAD~\cite{DDAD}~       & \uline{87.3}         & \uline{86.4}          & 74.3                  & 71.5                  & \uline{69.2}         & \uline{71.7}         & 84.5                  & 83.3                  & 75.3                  & 75.1                  & 78.1    & 77.6 \\
        & Multi-ST~\cite{MR-AD}~   & 86.0                  & 83.5                   & 68.1                  & 68.2                  & 60.8                  & 63.6                  & 95.6                  & \uline{92.7}         & 79.1                  & 74.4                  & 77.9    & 76.5 \\
        & RDAD~\cite{RD-AD}~       & 85.7                  & 82.9                   & 69.4                  & 66.3                  & 55.3                  & 56.9                  & 96.0                  & 92.5                  & 82.7                  & 78.5                  & 77.8    & 75.4 \\
        & Destseg~\cite{DeSTSeg}~    & 73.3                  & 73.9                   & 64.4                  & 66.8                  & 55.8                  & 56.0                  & \textbf{96.7} & \textbf{95.6} & 73.6                  & 72.1                  & 72.8    & 72.9 \\
        & Ours  (AE)  & 84.1                  & 82.4                   & \uline{76.6}         & 74.4                  & 65.2                  & 66.4                  & 89.2                  & 83.0                  & 82.5                  & \uline{79.0}         & 79.5    & 77.0 \\
        & Ours  (AEU) & \textbf{88.6} & \textbf{86.8}  & \textbf{78.7} & \textbf{76.1} & \textbf{72.9} & \textbf{71.8} & \uline{96.2}         & 92.0                  & \textbf{86.3} & \textbf{84.0} & \textbf{84.5}    & \textbf{82.6} \\ 
\bottomrule
\end{tabular}}
\end{table}

\begin{table}[t]
\centering
\caption{Ablation study for D2UE, where ``Ens" and ``Unc" stand for ``ensemble reconstruction score" and ``ensemble uncertainty estimation from output space", respectively. Results are in AUC and the best result is bold in each column.}
\label{tab:ablation}
\renewcommand\arraystretch{1.4}
\setlength{\tabcolsep}{4pt}{
\resizebox{\columnwidth}{!}{
\begin{tabular}{cccccccccccccccccccccc}
\toprule
\multicolumn{4}{c}{Methods} & \multicolumn{3}{c}{RSNA} & \multicolumn{3}{c}{VinDr-CXR} & \multicolumn{3}{c}{CXAD} & \multicolumn{3}{c}{Brain MRI} & \multicolumn{3}{c}{LAG} & \multicolumn{3}{c}{Average} \\
\cmidrule(lr){1-4} \cmidrule(lr){5-7} \cmidrule(lr){8-10} \cmidrule(lr){11-13} \cmidrule(lr){14-16} \cmidrule(lr){17-19} \cmidrule(lr){20-22} 
Ens                       & Unc                       & RAR                       & DSU                       & AE            & MemAE         & AEU           & AE            & MemAE         & AEU           & AE            & MemAE         & AEU           & AE            & MemAE         & AEU           & AE            & MemAE         & AEU           & AE      & MemAE & AEU  \\
\midrule
\Checkmark &                           &                           &                           & 66.9          & 68.0          & 86.7          & 55.9          & 55.8          & 73.8          & 55.6          & 56.0          & 66.4          & 79.7          & 77.4          & 94.0          & 79.3          & 78.5          & 81.3          & 67.5    & 67.1  & 80.4 \\
\Checkmark & \Checkmark &                           &                           & 69.4          & 68.0          & 87.3          & 60.1          & 59.5          & 74.3          & 59.8          & 59.4          & 69.2          & 59.8          & 52.6          & 84.5          & 72.1          & 70.4          & 75.3          & 64.2    & 62.0  & 78.1 \\
\Checkmark & \Checkmark & \Checkmark &                           & 76.8          & 77.9          & 87.8          & 71.8          & 71.6          & 75.6          & 62.9          & 62.4          & 72.4          & 61.1          & 63.9          & 91.0          & 79.6          & 78.7          & 79.9          & 70.4    & 70.9  & 81.3 \\
\Checkmark & \Checkmark &                           & \Checkmark & 80.8          & 81.6          & 88.5          & 68.5          & 66.2          & 77.5          & 61.1          & 62.6          & 72.6          & 88.9          & 86.2          & 95.1          & 82.4          & 78.5          & 83.7          & 75.5    & 75.0  & 83.5 \\
\Checkmark & \Checkmark & \Checkmark & \Checkmark & \textbf{84.1} & \textbf{83.5} & \textbf{88.6} & \textbf{76.6} & \textbf{75.7} & \textbf{78.7} & \textbf{65.2} & \textbf{63.1} & \textbf{72.9} & \textbf{89.2} & \textbf{87.5} & \textbf{96.2} & \textbf{82.5} & \textbf{79.0} & \textbf{86.3} & \textbf{79.5}    & \textbf{77.8}  & \textbf{84.5 } \\
\bottomrule
\end{tabular}}}
\end{table}

\noindent\textbf{Comparisons with State-of-the-Art Methods.} Table~\ref{tab:comparison} presents a comparison of our approach with an extensive assortment of state-of-the-art (SOTA) methods in AUC\% and AP\%. Compared SOTA methods include reconstruction-based methods such as MemAE~\cite{memAE}, AEU~\cite{AEU}, IGD~\cite{IGD}, f-AnoGAN~\cite{DBLP:journals/mia/fanogan}, Ganomaly~\cite{Ganomaly}, as well as ensemble uncertainty estimation methods such as DDAD~\cite{DDAD}, Multi-ST~\cite{MR-AD}, RDAD~\cite{RD-AD}, Destseg~\cite{DeSTSeg}.
Employing the AEU backbone, our method exhibits exceptional performance in comparison to other methods across multiple medical image datasets, such as RSNA, VinDr-CXR, CXAD, and LAG, in addition to achieving the second-highest AUC in Brain MRI. Specifically, it surpasses SOTA results in AUC and AP by 1.3\% and 0.4\% (RSNA), 2.1\% and 1.3\% (VinDr-CXR), 3.7\% and 0.1\% (CXAD), and 2.1\% and 5.0\% (LAG), demonstrating our method's effectiveness and superiority.

\begin{table}[t]
\centering
\caption{From left to right: No constraint, Euclidean and Manhattan distance (We use $e^{-\text{distance}(P,Q)}$ to stabilize training), Cosine similarity, Pearson correlation coefficient, CKA similarity. \Checkmark { }indicates the presence of the given mathematical property, while \XSolidBrush  { }indicates the opposite. The results are based on vanilla Autoencoder in the RSNA.}
\renewcommand\arraystretch{1.4}
\setlength{\tabcolsep}{4pt}{
\resizebox{\columnwidth}{!}{
\begin{tabular}{cccccccc}
\toprule
\multicolumn{2}{c}{Similarity metrics} & None       & Euclidean                   & Manhattan                   & Cosine                      & Pearson                     & CKA                       \\
\midrule
\multirow{2}{*}{Invariant to:}      & Isotropic scaling    & \textbf{—} & \XSolidBrush & \XSolidBrush & \Checkmark   & \Checkmark   & \Checkmark \\
                   & Orthogonal transform & \textbf{—} & \XSolidBrush & \XSolidBrush & \XSolidBrush & \XSolidBrush & \Checkmark \\
\midrule                
\multicolumn{2}{c}{AUC}              & 69.4       & 70.2                        & 71.3                        & 72.1                        & 72.9                        & 76.8 \\                      
\bottomrule
\end{tabular}
}
}
\label{tab:similarities}
\end{table}
\begin{figure}[!t]
  \centering
  \includegraphics[width=0.8\textwidth]{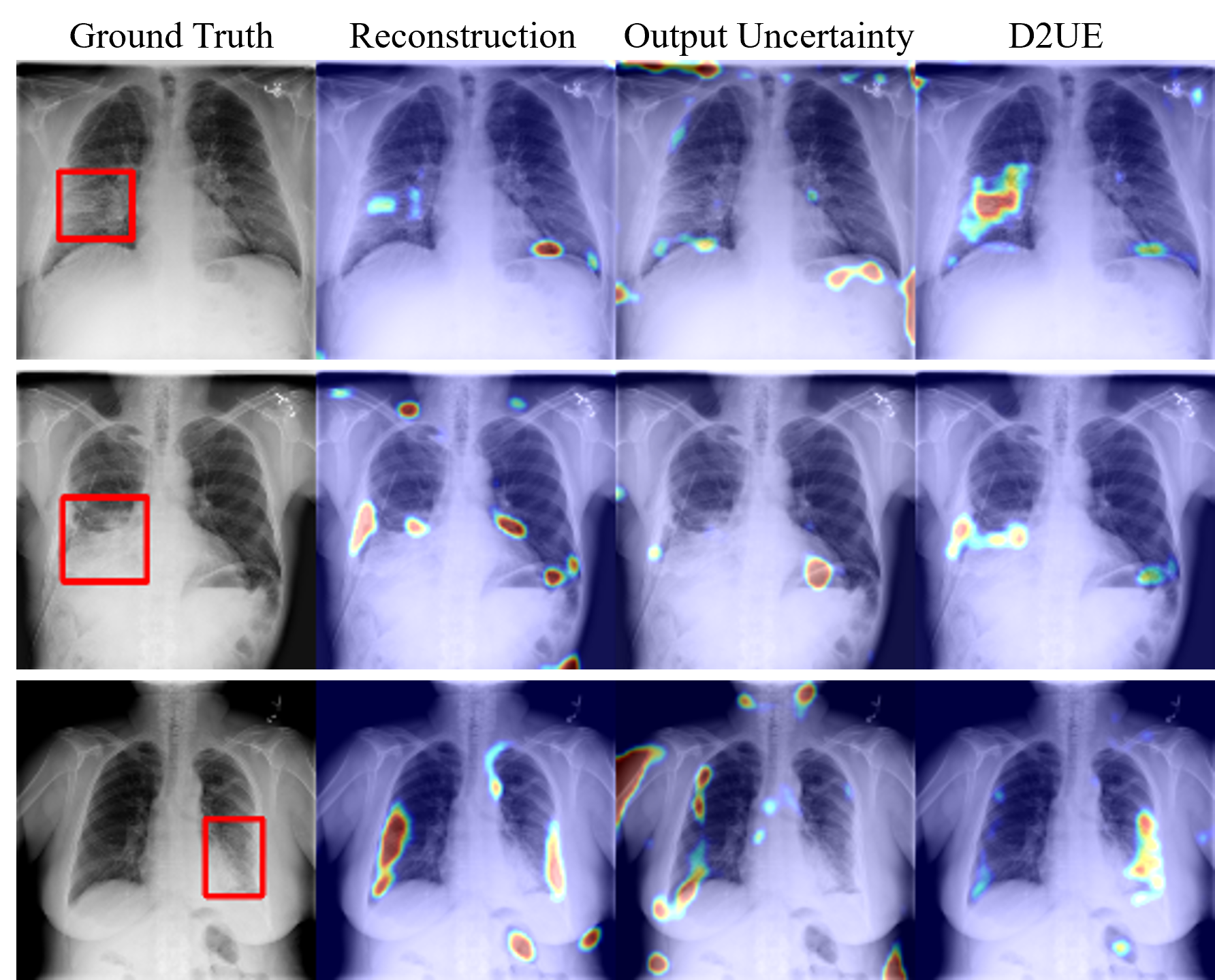}
  \caption{Visualization results. \textcolor{red}{Red} bounding boxes indicate abnormal regions.}
  \label{fig:visualization}
\end{figure}

\noindent\textbf{Ablation Study.}
\label{sec:ablation study}
We conduct an ablation analysis utilizing three distinct Autoencoder architecture backbones: AE, MemAE, and AEU, to scrutinize the effectiveness of each constituent of D2UE. Table~\ref{tab:ablation} showcases a comparison of the AUC\% for D2UE variations across five datasets. Variations include: 1) Ens, ensemble reconstruction score; 2) +Unc, ensemble uncertainty estimation from output space; 3) +RAR, incorporating RAR into training; 4) +DSU, utilizing DSU during inference. Results confirm that proposed components contribute positively to the enhancement of accuracy. For instance, on the RSNA, our RAR and DSU respectively improve the performance by 7.4\% AUC and 11.4\% within the AE model, demonstrating the effectiveness of our method. 

\noindent\textbf{Choice for similarity metric.}
\label{sec:similarity functions}
We examine various similarity metrics for RAR, and show results in Table.~\ref{tab:similarities}. It is discerned that the CKA similarity attains the highest AUC with 76.8\% among all other similarity functions. This is followed by the Pearson correlation coefficient (72.6\%), Cosine similarity (72.1\%), Manhattan distance(71.3\%), Euclidean distance (70.2\%), and no constraint (69.4\%). The empirical results substantiate that both scaling invariance and orthogonal invariance contribute positively to the accuracy.

\noindent\textbf{Visualization results.}
We visualize heat maps of ensemble reconstruction, ensemble uncertainty estimation from output space, and D2UE on RSNA in Fig.~\ref{fig:visualization}. Our method can significantly emphasize abnormal regions.  

\label{experiment}
\section{Conclusion}
In this paper, we presented D2UE, a Diversified Dual-space Uncertainty Estimation framework for medical anomaly detection.
To effectively balance the diversity among ensemble learners and reconstruction accuracy, we introduced redundancy-aware repulsion, which compels learners to disagree on anomalies without compromising agreement on normal inputs. 
Further, we propose dual-space uncertainty highlighting anomalous regions during inference to enhance the model's discrimination ability.
The framework has been extensively tested on various medical benchmarks, and experimental results demonstrate the superiority of our method to state-of-the-art methods and the effectiveness of each component. In future work, we intend to explore the quantitative relationship between ensemble diversity and final performance, and reduce the time and computational cost of training model ensembles.

\subsubsection{\ackname}
This work was supported by the Hong Kong Innovation and Technology Fund (Project No. MHP/002/22), Project of Hetao Shenzhen-Hong Kong Science and Technology Innovation Cooperation Zone (HZQB-KCZYB-2020083) and the Research Grants Council of the Hong Kong (Project Reference Number: T45-401/22-N).

\subsubsection{Disclosure of Interests.}
The authors have no competing interests to declare that are
relevant to the content of this paper.
\bibliography{Paper-1356}
\bibliographystyle{splncs04}
\end{document}